\documentclass{article} 
\usepackage[preprint]{colm2025_conference}

\usepackage{microtype}

\usepackage{inconsolata}

\usepackage{relsize}
\usepackage{amsmath}
\usepackage{amssymb}
\usepackage{mathtools}
\usepackage{amsthm}
\usepackage{hyperref}
\usepackage{url}
\usepackage{booktabs} 
\usepackage{caption}
\usepackage{subfigure}
\usepackage{graphicx}
\usepackage{amsthm}
\usepackage{float} 
\usepackage{bbm}
\usepackage{bm}
\usepackage{enumerate}

\usepackage{lineno}

\definecolor{darkblue}{rgb}{0, 0, 0.5}
\hypersetup{colorlinks=true, citecolor=darkblue, linkcolor=darkblue, urlcolor=darkblue}

\title{Identifying and Mitigating the Influence of the Prior \\ Distribution in Large Language Models}

\author{Liyi Zhang \\
Department of Computer Science\\
Princeton University\\
Princeton, NJ 08540, USA \\
\texttt{zhang.liyi@princeton.edu} \\
\And
Veniamin Veselovsky \\
Department of Computer Science\\
Princeton University\\
Princeton, NJ 08540, USA \\
\texttt{veniamin@princeton.edu} \\
\And
R.~Thomas McCoy \\
Department of Linguistics\\
Yale University\\
New Haven, CT 06511, USA \\
\texttt{tom.mccoy@yale.edu} \\
\And
Thomas L.~Griffiths \\
Department of Psychology and Computer Science\\
Princeton University\\
Princeton, NJ 08540, USA \\
\texttt{tomg@princeton.edu}}

\begin{document}

\ifcolmsubmission
\linenumbers
\fi

\maketitle

\begin{abstract}
Large language models (LLMs) sometimes fail to respond appropriately to deterministic tasks -- such as counting or forming acronyms -- because the implicit prior distribution they have learned over sequences of tokens influences their responses. In this work, we show that, in at least some cases, LLMs actually compute the information needed to perform these tasks correctly, and we identify some interventions that can allow them to access this information to improve their performance. First, we show that simply prompting the language model to not rely on its prior knowledge leads to dramatic improvements in prior-dominated tasks. We then use mechanistic interpretability techniques to localize the prior within the LLM and manipulate the extent to which that prior influences its responses. Specifically, we show that it is possible to identify layers of the underlying neural network that correlate with the prior probability of a response and that lightweight finetuning of these layers with basic prompts on prior-dominated tasks achieves high performance on held-out answers. These results suggest that the information required to produce a correct response is contained within the representations of the problems formed by the models. Furthermore, we show that this finetuning is significantly more effective for prior-dominated tasks, and that the error after finetuning is no longer correlated with the prior. Our results suggest that it may be possible to define effective methods for manipulating the extent to which LLMs rely upon their priors in solving problems, potentially increasing their performance in settings where LLMs hallucinate for reasons related to the prior probability of token sequences.
\end{abstract}

\section{Introduction}

\begin{figure}[t]
    \centering
    \subfigure[Shift-cipher - prompt-response example]{
    \includegraphics[width=0.85\linewidth]{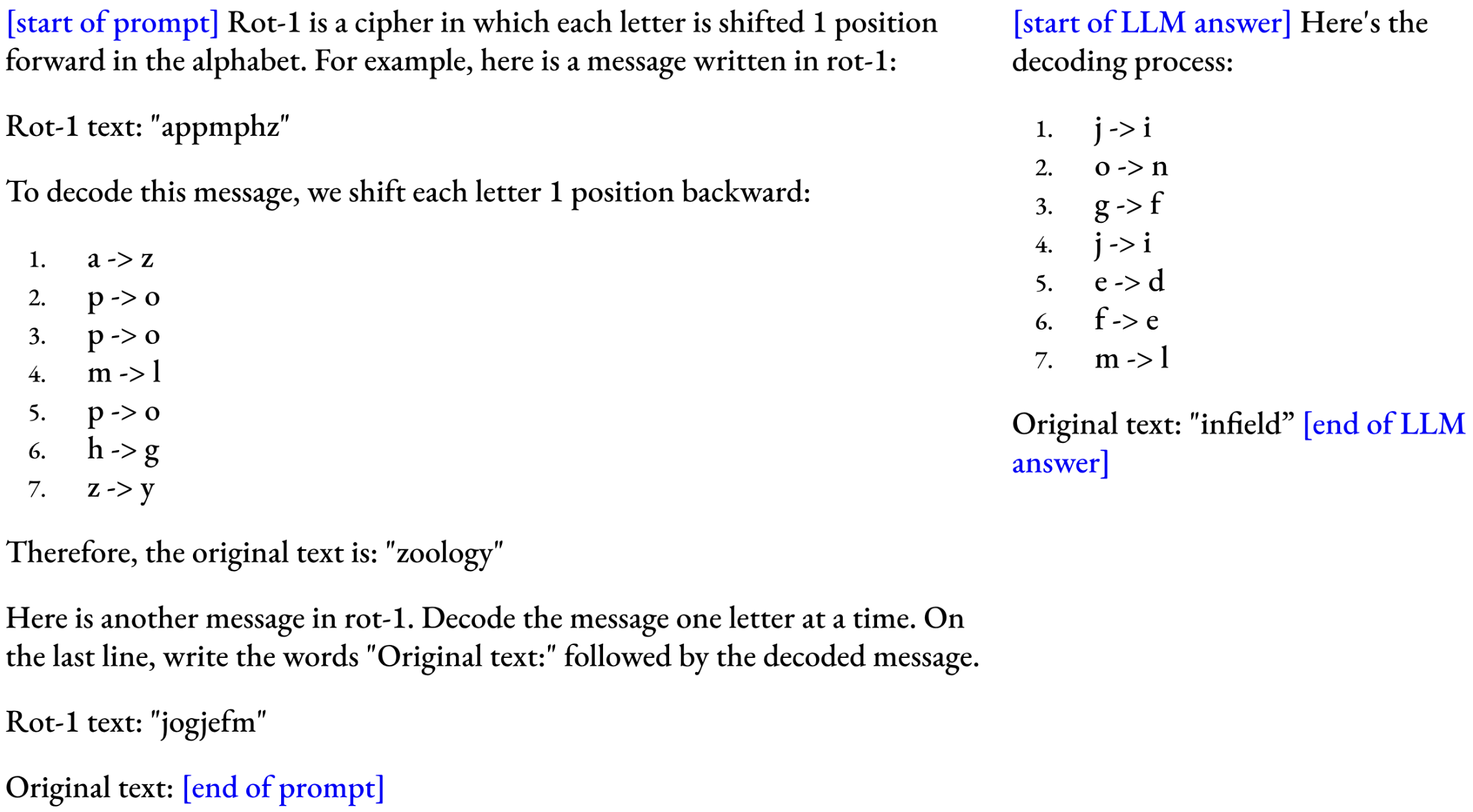}
    \label{fig:shift-cipher-example}}
    \subfigure[Pretrained model accuracy on three tasks]{
    \includegraphics[width=0.7\linewidth]{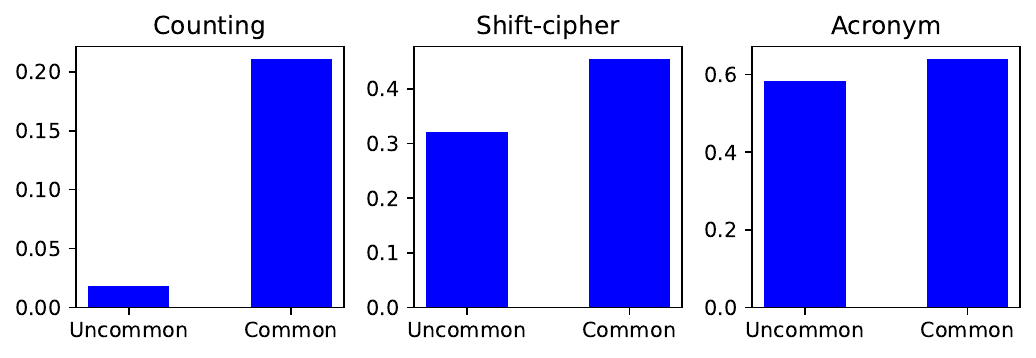}
    \label{fig:overall-base}}
    \caption{Influence of the prior on \textsc{Llama 3}'s performance for deterministic tasks. (a): Although \textsc{Llama 3} reasons through the correct characters, it outputs a more likely token `in', instead of the correct `inf' (which is the first token of the correct answer `infidel'), followed by `field'. (b): Performance of the pretrained LLM \textsc{Llama 3} on three prior-related tasks. For counting, the common set is multiples of 10 from 20 to 100; the uncommon set is all the other numbers from 11 to 100. For shift-cipher and acronym, we take two sets of answer vocabulary items whose words appear relatively commonly and uncommonly in natural texts.}
    \label{fig:counting-across-numbers}
\end{figure}

Large language models (LLMs) are capable of producing coherent text in a variety of settings \citep{radford2019language,dou2021scarecrow,bubeck2023sparks,chang2024language}, yet they often fail at simple tasks, producing hallucinations and having difficulty performing logical reasoning \citep{lin-etal-2022-truthfulqa, mccoy2023embersautoregressionunderstandinglarge,wu2024reasoning,mirzadeh2024gsm,razeghi2022impact,stechly2024chain}. \citet{mccoy2023embersautoregressionunderstandinglarge} argue that these failures are partly a consequence of LLMs having difficulty producing low-probability output. For example, when solving puzzles such as deciphering a message by shifting each letter in the message by one position in the alphabet, LLMs will perform better when the correct answer is a high-probability string than when it is a low-probability string, even though the underlying logic of these tasks is same (Figure \ref{fig:shift-cipher-example} provides an example of one of these errors being reproduced by \textsc{Llama 3}). 
One way to understand these errors is to assume that LLMs perform Bayesian inference \citep{griffiths2024bayes}, letting the prior distribution over word sequences that they have learned through pre-training on large amounts of text influence their output \citep{mccoy2023embersautoregressionunderstandinglarge}. However, it is unclear what the mechanisms behind this influence might be, and whether their effects can be mitigated.


Identifying the mechanisms underlying the behavior of LLMs is challenging. 
LLM residual streams (i.e.~the internal representations produced by processing sequences of tokens) are known to be \textit{polysemantic}, where each hidden unit can be activated by multiple different concepts \citep{templeton2024scaling,Cunningham2023SparseAF} due to the way in which neural network representations naturally instantiate superpositions of multiple pieces of information \citep{smolensky1986neural}. Creating probes -- neural networks that are trained to predict information of interest based on the internal representations of the models \citep{ettinger2016probing,adi2017fine,hupkes2018visualisation,belinkov2021probing}  -- is a common approach for evaluating the implicit knowledge of LLMs \citep{nanda-etal-2023-emergent,gurnee2024languagemodelsrepresentspace}. For example,
\citet{orgad2024llmsknowshowintrinsic} showed that LLMs know more than they show by conducting probing experiments on multiple choice questions. However, probing can confound what is embedded inside LLMs with what is learned by the probe \citep{wieting2018no,hewitt2019designing}. This has led to the development of methods for mechanistic interpretability, such as logit-lens \citep{Belrose2023ElicitingLP} or using interventions on the values of hidden units to evaluate their causal effects \citep{giulianelli2018hood,soulos2020discovering,geiger2021causal,minder2025controllablecontextsensitivityknob}.


In this paper, we show that the implicit prior learned by LLMs results in incorrect responses even though the model has internal representations that are sufficient to solve the problem. We identify potential sites for intervention  by localizing the prior within the LLM. Analyzing individual layers in \textsc{Llama 3} \citep{grattafiori2024llama3herdmodels} with logit lens, we find that layers tend to either have a strong or no correlation with the prior. This suggests that the prior is encoded in the residual stream, but only at certain levels of the model. To find information leading to the correct answer that is potentially embedded in the LLMs, we explore two methods: adding a simple prompt in-context, and finetuning on a stratified train-validation setup. First, we find that simply adding ``do not rely on your prior knowledge'' to the prompt significantly improves performance across some prior-dominated tasks. However, the stratified finetuning method achieves even more consistent performance. 
Comparing the results of stratified finetuning across different tasks shows that its benefits are greatest for tasks where the prior influences behavior. This suggests that we are effectively eliciting knowledge already encoded in LLMs to improve performance in settings where the prior steers the model away from the correct answer. 

The performance increase from these methods suggests that the ability to solve the task already exists in the LLM, but the model is led away from these answers by its prior. We also show that for generative tasks where the LLM must output a word or a number, the probe itself can learn to perform the task, and thus that a stratified setup is needed to explore this problem. Finally, we demonstrate that the methods we have developed are appropriately targeted at reducing the influence of the prior in settings where the prior is not helpful: we show that the finetuned improvement in performance is significantly smaller for  tasks where the prior is not problematic, and that finetuned performance no longer correlates with the probability of the answers but rather with the difficulty of the questions. 

Our main contributions are revealing how information about prior distributions and the task at hand is encoded by LLMs, and defining simple yet effective methods for manipulating the extent to which LLMs rely upon these priors, increasing their performance in settings where the LLM can already solve problems if the answers are high-probability tokens. More specifically:
\begin{enumerate}[1.]
    \item We find specific locations inside LLMs from which the prior probability over tokens can be decoded.
    \item We show that, in several deterministic tasks where LLMs show probability sensitivity, the residual stream encodes information needed to perform the task, suggesting that the incorrect responses can be attributed to the LLM getting distracted by the prior.
    \item We identify two approaches -- one based on prompting and one based on fine-tuning -- that are effective for intervening on the use of prior knowledge by LLMs, enabling their behavior to better reflect the information they contain that is relevant to the task.
\end{enumerate}

\section{Related Work}



\subsection{Hallucination in LLMs}


LLMs sometimes output incorrect content driven by a variety of factors including biases \citep{kotek2023gender}, factual inaccuracy \citep{lin-etal-2022-truthfulqa,liu2022token}, and errors in reasoning \citep{wu2024reasoning,mirzadeh2024gsm,zhu2024incoherentprobabilityjudgmentslarge}. These various types of errors are sometimes grouped together under the term \textit{hallucination}. Note that there is inconsistency in the field in how the term \textit{hallucination} is used \citep{venkit2024audit}; we follow \citet{orgad2024llmsknowshowintrinsic} in using it to mean any type of error produced by LLMs. 

Our work explores one type of hallucination, where an LLM prefers to output higher-probability sequences of tokens in settings where the correct response has lower probability. \citet{mccoy2023embersautoregressionunderstandinglarge} documented this phenomenon via extensive analysis of the behavior of LLMs. \citet{Prabhakar2024DecipheringTF} extended this work by using chain-of-thought prompting to show that LLM behavior involves a mixture of memorization and reasoning on shift-cipher problems. Our work builds on \citet{mccoy2023embersautoregressionunderstandinglarge} from a different angle, where we focus on localizing the influence of the prior and developing novel methods for removing its effect on a range of prior-dominated problems. \citet{orgad2024llmsknowshowintrinsic} reached a similar conclusion as ours -- that LLMs know more than they show -- but by focusing on problems that allow free responses (as opposed to the multiple choice questions that were the focus of their analysis), we generalize these results to a wider range of settings. 

\subsection{Mechanistic understanding of the effect of priors on text generation}

Existing work has explored manipulating the impact of learned prior probabilities on text generation. \citet{minder2025controllablecontextsensitivityknob} did this by identifying a one-dimensional subspace that controls whether text generation samples from the prior or from the context provided to the model. Other work has explored how an LLM functions under uncertainty and found that there exist two mechanisms that encode uncertainty: the entropy neuron~\citep{gurnee2024universal,katz-belinkov-2023-visit} and pushing generation towards the unigram prior~\citep{stolfo2024confidenceregulationneuronslanguage}. This ability to push the LLM generation towards the unigram prior has been shown to be an effective strategy for controlling text generation~\citep{nielsen2025predictionhubscontextinformedfrequent}. 

\subsection{Controlling text generation}

The ability of the user to control LLM text generation in general has been extensively researched. In this paper, we take inspiration from prompt engineering, finetuning, probing, and steering. Prompting has been shown to be able to control text generation through personas~\citep{Hu2024QuantifyingTP}, chain-of-thought reasoning~\citep{Wei2022ChainOT, Yao2023TreeOT}, refinement, and more esoteric techniques~\citep{salinas2024butterfly}. Other approaches focus on parameter updates including full finetuning~\citep{bommasani2021opportunities} and parameter efficent finetuning (PEFT) techniques such as like LoRA or ReFT~\citep{Hu2021LoRALA, Wu2024ReFTRF}. Another line of work controls the model through steering techniques such as dictionary learning~\citep{Cunningham2023SparseAF}, contrastive activation addition~\citep{Rimsky2023SteeringL2}, and distributed alignment search~\citep{Geiger2023FindingAB,minder2025controllablecontextsensitivityknob}. 





\section{Approach}

\subsection{Model}

For all tasks, the LLM that we evaluate is the instruction-tuned \textsc{Llama 3} \citep{grattafiori2024llama3herdmodels} with 8B parameters.
We chose \textsc{Llama 3} because it satisfies the two criteria that are necessary for our analyses. First,
its weights are open, which is a prerequisite for our goal of analyzing the model's internal processing. Second, it is capable of achieving reasonably high performance on the tasks that we use (\citet{mccoy2023embersautoregressionunderstandinglarge} find that some open-weights models achieve scores of 0\% on these tasks, preventing meaningful conclusions from being drawn from them).

\subsection{Tasks}

Our experiments focus on two kinds of tasks: tasks in which the prior tends to result in incorrect responses (``prior-dominated tasks'') and tasks in which the prior seems to have little effect (``prior-insensitive tasks''). We describe these two kinds of tasks in turn. Except for the ``make letters'' task, similar versions of all the tasks can also be found in \citet{mccoy2023embersautoregressionunderstandinglarge}.

\subsubsection{Prior-dominated tasks}


\paragraph{Counting.} The LLM is presented with a sequence of identical letters (e.g. 23 ``m''s), and then it is asked how many letters are there in this list (e.g. 23). The LLM prefers to output numbers that are common, such as multiples of 10 (Figure~\ref{fig:counting-across-numbers}b, left).

\paragraph{Shift cipher.} The LLM is presented with a sequence of letters (e.g. ``bqqmf'') and is asked to shift each letter forward by $n$. For example, if $n=1$, the correct answer for ``bqqmf'' would be ``apple''. The LLM prefers to output common words (Figure~\ref{fig:counting-across-numbers}b, middle).

\paragraph{Acronyms.} The LLM is presented with a list of words (e.g. ``Counter Affairs Trigram'') and is asked to concatenate the first letter of each word (e.g. ``CAT''). The LLM prefers to output common words (Figure~\ref{fig:counting-across-numbers}b, right).


\subsubsection{Prior-insensitive tasks}


\paragraph{Multiplication.} The LLM is asked to multiply two three-digit positive integers. The answers for such queries are numbers that are so large that all of them are very rare in natural text; therefore, the LLM is unlikely to have strong priors over these output sequences, justifying our designation of this task as prior-insensitive.

\paragraph{Make letters.} This task can be considered as ``reverse counting'', where the prompt asks the LLM to form a number of letters (e.g. ``7 letter `c's separated by spaces''). The correct answer would be ``c c c c c c c''. As with the multiplication case, all correct answers for this task are such low-probability strings that the LLM is unlikely to have established strong priors over these strings, justifying our classification of this task as being largely prior-insensitive.


\subsection{LLM internal representations}

To describe the methods we use for exploring the internal representations of LLMs, we introduce some formal notation. We will denote an input sequence (such as the prompt used in our experiments) as $x_{1:T_1} = \{x_1, x_2, ..., x_{T_1}\}$. Given this input, the LLM produces the output sequence $x_{T_1+1:T_1 + T_2} = \{x_{T_1+1}, x_{T_1+2}, ..., x_{T_1 + T_2}\}$. In the transformer architecture \citep{Vaswani+2017}, each token $x_i$ in the input and output sequence has an associated embedding $z^{(L)}_i$ in the final layer $L$ of the model. These embeddings make up the residual stream. 
On lower layers $l$, we denote the token's corresponding embedding as $z^{(l)}_i$. Except for the last layer, $z^{(l)}_i$ is associated with $z^{(l-1)}_i$ by the attention mechanism across $i$.

\subsection{Mechanistic interpretability methods}


\subsubsection{In-context prompting}
Past work has demonstrated that LLMs can reason about their own generations \citep{tian-etal-2023-just,kadavath2022language} and that they can obey instructions about how to carry out a task, such as performing it step-by-step \citep{kojima2022large}. 
We hypothesize that LLMs may similarly capture meta-knowledge about their priors that could enable them to be guided to avoid relying on their priors when generating an answer.
To investigate this possibility, we simply append the string ``do not rely on your prior knowledge'' (or ``do not rely on your prior knowledge on the output'' in the case of shift ciphers) to the end of the prompt. 


\subsubsection{Probing} 

We explore the efficacy of probing methods on what LLMs know about the answers to a task -- a method that has previously been used for revealing knowledge about multiple choice questions \citep{orgad2024llmsknowshowintrinsic}. Probing aims to understand whether the target answer for a given task exists in the model's internal representations. An intuitive way to do so is to use a linear probe on LLM embeddings. In a given experiment, we arbitrarily select a fixed token $i$ and layer $l$. For each sequence in the dataset, the LLM produces internal embedding $z^{(l)}_i$. The probe is defined as a parameterized function $g = \text{Softmax}(\text{Linear}(z^{(l)}_i))$ mapping from the space of embedding $z^{(l)}_i$ to a probability vector representing the answer for the task.

Additional challenges exist in probing on word prediction tasks. If the probe targets the space of all LLM tokens, then the classes are sparse, and the probe would need to zero-shot predict classes during the validation stage. Thus, we insert the probe as a dense layer between an LLM hidden state and the LLM output layer. Similarly, for each sequence in the dataset, the LLM produces internal embedding $z^{(l)}_i$ at token $i$ and layer $l$. The probe is defined as a parameterized function $g = \text{ReLU}((\text{Linear})$) mapping from the embedding space $z^{(l)}_i$ to a vector representing the input to the LLM output layer.

\subsubsection{Finetuning}

Because the relationship of the prior to model embeddings is non-linear (which we will discuss in Section \ref{sec:other-techniques}), finetuning is potentially a more effective approach than probing for removing the  effect of the prior and bringing out knowledge encoded by the LLM. 

Unlike previous work, we use a stratified train-validation split to prevent influence from potential confounding variables and overfitting. In each task, it is ensured that all answer tokens used in the validation set do not appear in the training set. This ensures that the finetuned models do not memorize arbitrary relationships between the answer and patterns that appear in the question. For example, correctly answering that a sequence is 43 letters long means that the original model stores information about the concept of a 43-letters-long sequence, since the number 43 is unseen during training. A similar logic applies to piecing together letters into a word in other tasks.

The stratified setup is important in determining that the LLM encodes knowledge necessary to solve the task. As the results we report later in the paper show, probing can achieve high accuracy on a random train-validation split, but it gets close to $0\%$ accuracy on a stratified split.

\subsubsection{Linear intervention methods}
\label{sec:other-techniques}
In addition to the approaches outlined above, we also studied two additional techniques to reduce reliance on the prior motivated by previous work. In the first approach, we used the one-dimensional prior-context subspace from \citep{minder2025controllablecontextsensitivityknob}. In the second approach, we ablate on the unigram latent direction found in \citep{stolfo2024confidenceregulationneuronslanguage}.  
Since these results did not lead to significant improvements, we relegate the results and explanations of methods to Appendix~\ref{sec:appendix-techniques}. 


\section{Empirical Evaluations}

\subsection{Localization of prior influence}

\paragraph{\textsc{Llama 3} prefers common answers.} In this paragraph, we illustrate a phenomenon that is reproduced from \citet{mccoy2023embersautoregressionunderstandinglarge}. Figure \ref{fig:counting-across-numbers-original} shows that on the counting task, \textsc{Llama 3} prefers to output answers that are common, which in this case is multiples of ten. In particular, \textsc{Llama 3} achieves high accuracy for sequence lengths that are common numbers, and yet has $0\%$ accuracy on most other sequence lengths. We also find that most numbers are never outputted as answers, although the lengths of sequences in our questions are evenly distributed across all integers from 1 to 100. Figure \ref{fig:overall-base} shows that similar probability sensitivity arises in the other tasks as well.

Since this result suggests that \textsc{Llama 3} is capable of counting certain sequences, we explore which parts of the model steer its output away from the correct answer but instead towards common tokens, and whether we can target the removal of this prior influence and achieve high performance.

\begin{figure}[t]
    \centering
    \includegraphics[width=0.5\linewidth]{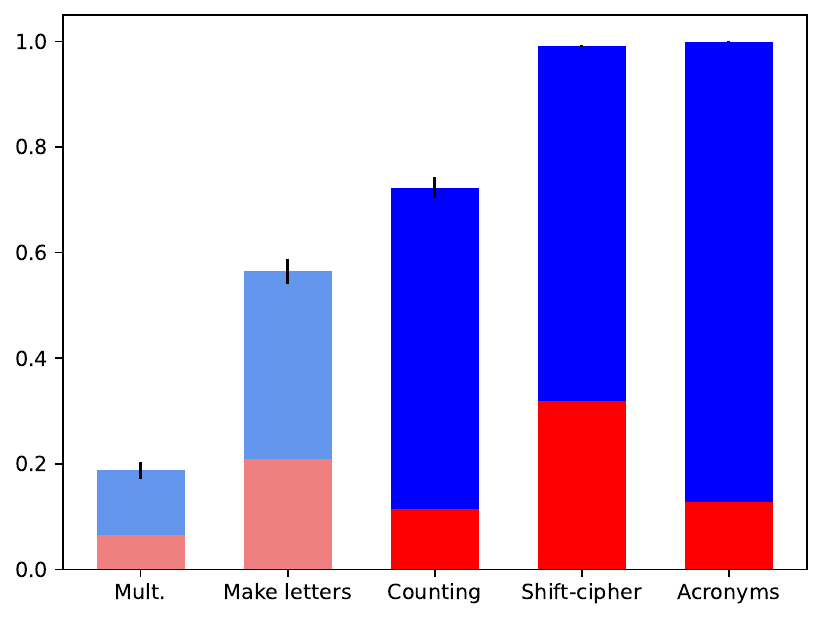}
    \caption{Accuracy of finetuned (blue) and original (red) models on six tasks (shown here is the best layer's performance in each individual task). Of these tasks, the multiplication and make letters task involve little prior influence. This suggests that finetuning on a specific task is more effective for tasks where the prior steers the model away from the right answer.}
    \label{fig:overall-ft}
\end{figure}

\paragraph{LLM layers exhibit an all-or-none pattern in encoding the prior.} We use logit lens to investigate which individual layers embed the prior. To illustrate how the analysis of where the prior is embedded is constructed, we use the example of the shift-cipher task. Other tasks are investigated similarly. The construction of this analysis can be broken down into three steps:
\begin{enumerate}[1.]
    \item \textbf{Dataset of tokens.} We use a dataset of two-token English words that are seven characters long. These serve as true answers to the shift-1 cipher questions. Logits that will be acquired later are recorded only for the first tokens of each of these words.
    \item \textbf{Prompt construction.} \textsc{Llama 3} answers a shift-1 question with a prompt that is similar to the one we used previously to evaluate the LLM on the shift cipher, except that the prompt tells the LLM that it should attempt to guess the answer before the input (the encoded text) has been provided. This framing aims to encourage the LLM to fall back on its prior over possible answers because it has no other information that it can rely on. This prompt follows the actual question-answer prompts as closely as possible by, for example, making sure that there is no hint that the answer is an English word. 
    \item \textbf{Acquiring logits.} Finally, we record the logits for the relevant tokens gathered in the first step. The logits are \textsc{Llama 3}'s output at the token where it makes its answer to the prompt in the above step.
\end{enumerate}

To inspect what information about the answer appears in individual layers, we employ logit lens. For each layer $l$, we acquire a list of LLM embeddings over its answer to each shift-cipher question, $z_i^{(l)}$, where $i$ denotes the index of the output token corresponding to the answer position. We multiply each layer's embedding with the unembedding layer to acquire a proxy of logits over the LLM answer. We compute the Spearman correlation between the rankings of the LLM prior logits (obtained as described in steps 1 through 3 above) and the LLM answer logits (obtained via logit lens applied to the LLM's hidden state on standard prompts in which the input text has been provided, rather than requiring the LLM to guess without the input provided). For example, using a dataset of 10,000 words corresponding to 10,000 shift-cipher questions, we get 10,000 correlations. 

\begin{figure*}[t]
    \centering
    \subfigure[Counting]{
    \includegraphics[width=0.31\linewidth]{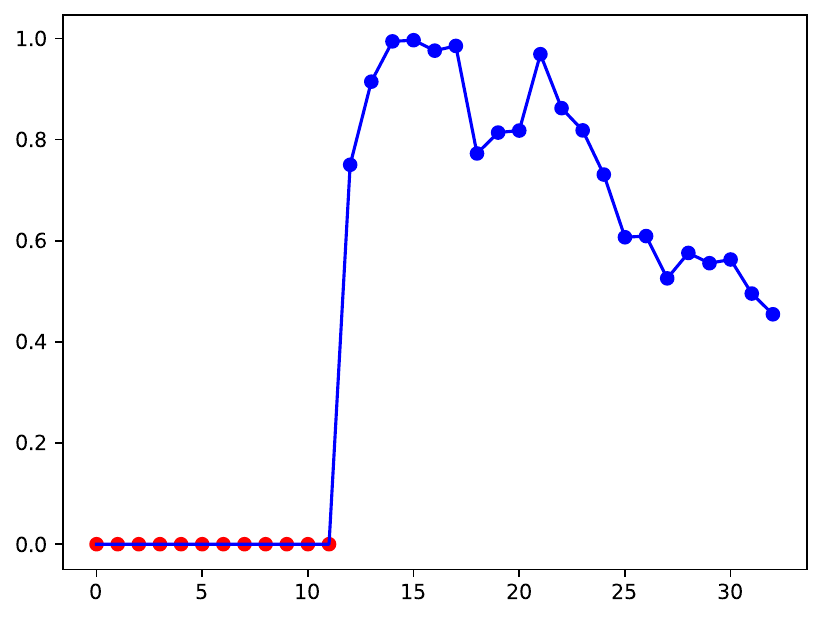}
    }
    \hfill
    \subfigure[Shift-cipher.]{
    \includegraphics[width=0.31\linewidth]{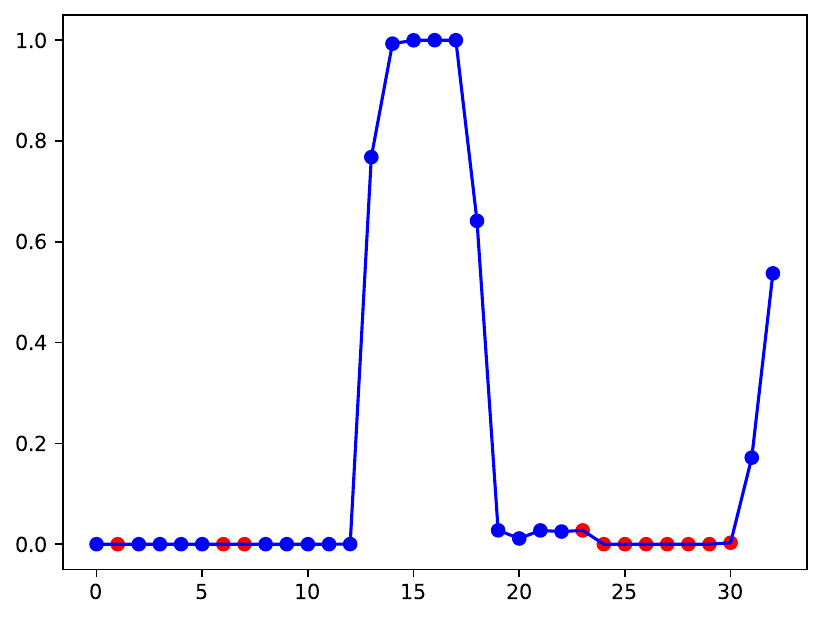}
    }
    \hfill
    \subfigure[Acronyms.]{
    \includegraphics[width=0.31\linewidth]{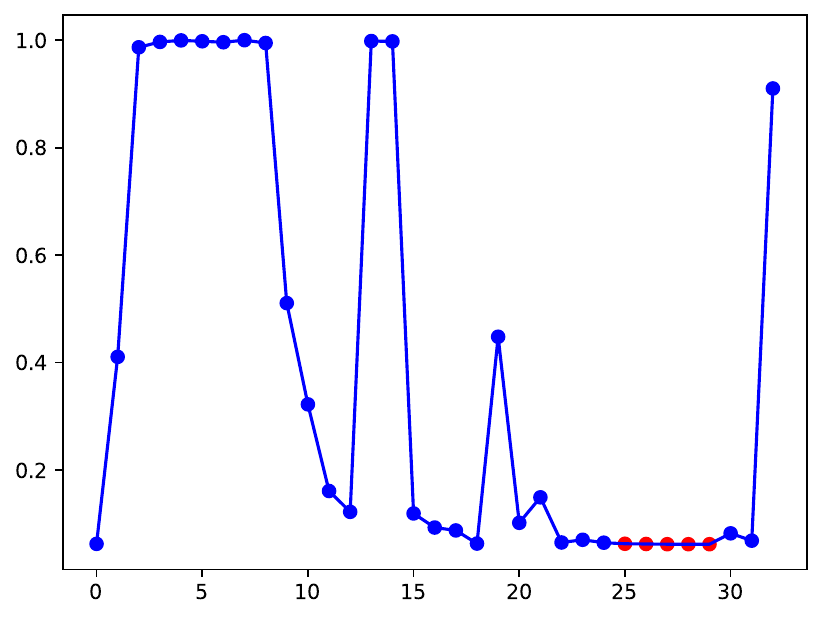}
    }
    \caption{Percentage of questions (vertical axis) where the LLM answer logits have a positive correlation with the prior with p-value $<0.05$, versus LLM layer number (horizontal axis) in an LLM with 32 total layers. Dots colored in red have more answer logits with negative correlation. \textit{Higher implies stronger correlation.}}
    \label{fig:corr}
\end{figure*}

Results for three tasks are shown in Figure \ref{fig:corr}. The encoding of the prior is not evenly distributed: most layers either exhibit significant correlation on all questions, or on no questions at all. This pattern suggests that some layers do not focus on encoding the prior, which further suggests that one might be able to delineate knowledge of the true answer from prior knowledge.

\subsection{Using information from internal representations}

The previous results suggest that LLMs encode the correct answers on tasks even when they are incorrectly influenced by the prior. This suggests that it should be possible to develop methods for extracting those correct answers. We did this in three ways: in-context prompting, probing, and finetuning. We found that probing generalized poorly but the other two methods were effective.

\paragraph{In-context prompting.} Table \ref{tab:ft} shows that in-context prompting achieves significant improvement in two of the three tasks: counting and acronyms. Because we can be sure that the LLM receives no extra information with this extra line of the prompt (other than the encouragement to avoid relying on the prior), this result suggests that at least in some prior-dominated tasks, the LLM is capable of solving the problem but the prior steers its response away from the correct answer. A mild improvement is also observed on the shift-cipher task. 

\paragraph{Probing.} We conducted probing on the counting, shift-cipher, and acronyms tasks. While probing improves performance significantly when the train-validation sets are randomly split, validation accuracy on the stratified set steadily drops from random-guessing to $0\%$ during probe training. This suggests that probing on these generative tasks (where the model outputs words or numbers instead of a multiple-choice) does not reliably decode knowledge from the model, but is subject to memorization within the probe.

\paragraph{Finetuning.} We used low-rank adaptation \citep[LoRA; ][]{Hu2021LoRALA} in finetuning (for further implementational details, see Appendix \ref{sec:appendix-implementation}). Figure \ref{fig:overall-ft} and Table \ref{tab:ft} show substantial improvement achieved by this lightweight finetuning across all prior-dominated tasks. All experiments are done in the stratified setup, so no validation answers are seen during finetuning. 
Because our previous evidence suggests that the relevant knowledge is already encoded within the LLM, lightweight finetuning on only 2000 to 8000 sequences with 50 epochs suffices to achieve high performance consistently across tasks. Just as probing fails in the stratified setup, finetuning the whole model can also be affected by overfitting as seen on the counting task in Table~\ref{tab:ft}. However, finetuning only one layer in the earlier layers consistently achieves higher performance and mitigates the overfitting issue.

\begin{table}[t]
    \centering
    \caption{Validation accuracy with standard error for the original model (which is the standard version of \textsc{Llama 3}), finetuned models, and the original model with a prompt that discourages reliance on the prior. Finetune-layer refers to performance obtained by only finetuning the best-performing layer. }
    \resizebox{0.65\columnwidth}{!}{
    \begin{tabular}{ccccc}
    \toprule 
    Task & Counting & Shift-cipher & Acronym\\
    \midrule
    Original & $11.5\% \mathsmaller{\;\pm 1.4\%}$ & $31.9\% \mathsmaller{\;\pm 1.0\%}$ & $12.8\% \mathsmaller{\;\pm 1.1\%}$\\
    Finetune & $37.5\% \mathsmaller{\;\pm 2.1\%}$ & $99.2\% \mathsmaller{\;\pm 0.2\%}$ & $97.5\% \mathsmaller{\;\pm 0.5\%}$\\
    Finetune-layer & $72.3\% \mathsmaller{\;\pm 2.0\%}$ & $99.4\% \mathsmaller{\;\pm 0.2\%}$ & $99.9\% \mathsmaller{\;\pm 0.1\%}$ \\
    Prompting & $49.2\% \mathsmaller{\;\pm 2.2\%}$ & $34.5\% \mathsmaller{\;\pm 1.1\%}$ & $46.3\% \mathsmaller{\;\pm 1.6\%}$ \\
    \bottomrule
\end{tabular}
    \label{tab:ft}}
\end{table}

\subsection{Finetuning works better on prior-dominated tasks}

\paragraph{Performance improvement differs significantly across tasks related and not related to the prior.} 
Figure \ref{fig:overall-ft} compares our three prior-dominated tasks to our two tasks that we have argued are unlikely to have significant prior influence. The performance increase is significantly higher for the prior-dominated tasks. This pattern remains the same if all tasks use whole-model finetuning instead of best-layer finetuning (with a similar drop in both the counting task and the make-letters tasks). This suggests that our lightweight finetuning is most useful when the prior knowledge over outputs can be expected to sway LLM responses. It also suggests that while even stratified finetuning can provide the model with some ability to perform the task (as seen by the performance improvement that it provides even in prior insensitive tasks), it also performs the function of prior removal (as seen by the fact that it provides larger benefits when we would expect prior removal to be helpful than when we would not expect it to be).
 
\begin{figure}[t]
    \centering
    \subfigure[Counting - original model]{
    \includegraphics[width=0.43\linewidth]{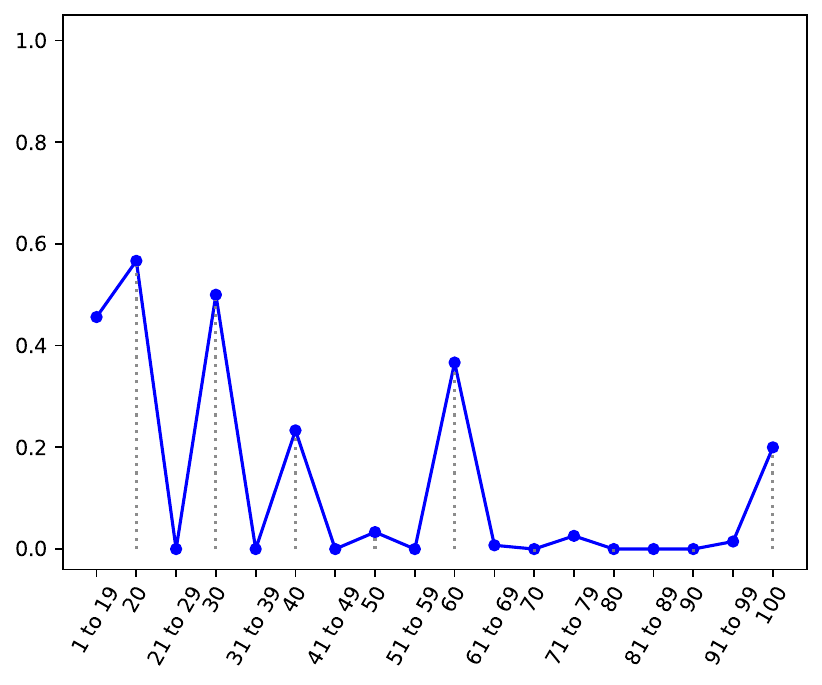}
    \label{fig:counting-across-numbers-original}}
    \subfigure[Counting - finetuned model]{
    \includegraphics[width=0.43\linewidth]{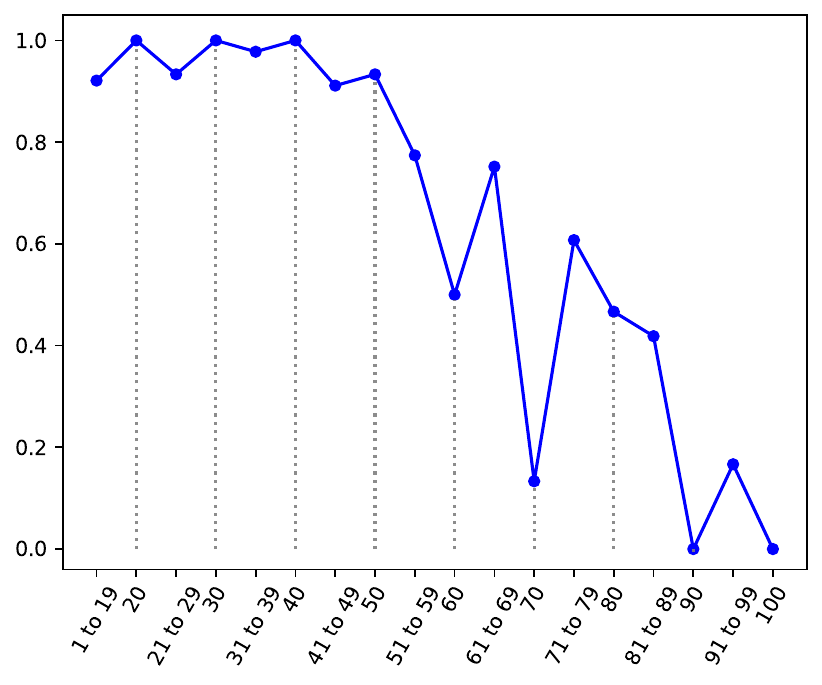}
    \label{fig:counting-across-numbers-ft}}
    \caption{Accuracy of \textsc{Llama 3} (y-axis) when asked to count different lengths of sequences of letters (x-axis). The original LLM performance is biased towards common numbers such as multiples of ten (marked by grey dash lines). The finetuned performance is instead correlated with lengths of sequences.}
\end{figure}

\paragraph{Finetuned answers do not show a preference towards common tokens.} Figure \ref{fig:counting-across-numbers-original} and \ref{fig:counting-across-numbers-ft} further support the view that the fine-tuning we have done serves the purpose of prior removal: post-finetuning performance on individual questions in the counting task correlates with the natural difficulty of the questions, where longer sequences have lower accuracy. There are no longer spikes at the common numbers, further showing that the bias towards prior knowledge is reduced with this approach.

\section{Conclusion}

In this work, we explore a set of problems where LLMs err but have strong potential of being correct on: prior-dominated problems where the LLM prefers to output common tokens even on deterministic tasks. It has been unknown whether the LLM encodes information needed to solve the task, or whether it simply relies on statistical heuristics because of its autoregressive maximum-likelihood training. Our work finds that the information necessary to solve the tasks we have analyzed is encoded by the model. We have also identified some challenges for the task of understanding the representation of the prior: the prior is encoded in the LLMs in a complex way such that linear methods such as steering do not improve performance, and probing cannot test the model's understanding on a stratified validation setup. Meanwhile, other computationally efficient methods show that the LLM possesses the information needed to solve the task. Prompting a language model to reduce its reliance on the prior can increase performance, but the amount of increase is inconsistent. Complementary to prompting, we also use lightweight finetuning on the language model, which can result in significantly improved performance across prior-dominated tasks in the stratified setup. We hope that our work lends insight into the nuanced interplay between the errors that LLMs make and the information that LLMs encode, and that these insights can lead to future improvements in situations where LLMs hallucinate for reasons related to the prior probability of token sequences.

\section*{Reproducibility Statement}

We use torchtune \citep{torchtune} for finetuning. Implementation details are provided in Appendix \ref{sec:appendix-implementation}. Code can be found at \href{https://github.com/zhang-liyi/llm-prior}{https://github.com/zhang-liyi/llm-prior} .


\section*{Acknowledgments}

This work was supported by ONR grant number N00014-23-1-2510.


\bibliography{colm2025_conference}

\begin{thebibliography}{53}
\providecommand{\natexlab}[1]{#1}
\providecommand{\url}[1]{\texttt{#1}}
\expandafter\ifx\csname urlstyle\endcsname\relax
  \providecommand{\doi}[1]{doi: #1}\else
  \providecommand{\doi}{doi: \begingroup \urlstyle{rm}\Url}\fi

\bibitem[Adi et~al.(2017)Adi, Kermany, Belinkov, Lavi, and Goldberg]{adi2017fine}
Yossi Adi, Einat Kermany, Yonatan Belinkov, Ofer Lavi, and Yoav Goldberg.
\newblock Fine-grained analysis of sentence embeddings using auxiliary prediction tasks.
\newblock In \emph{{International Conference on Learning Representations}}, 2017.

\bibitem[Belinkov(2021)]{belinkov2021probing}
Yonatan Belinkov.
\newblock Probing classifiers: Promises, shortcomings, and alternatives.
\newblock \emph{arXiv e-prints}, pp.\  arXiv--2102, 2021.

\bibitem[Belrose et~al.(2023)Belrose, Furman, Smith, Halawi, Ostrovsky, McKinney, Biderman, and Steinhardt]{Belrose2023ElicitingLP}
Nora Belrose, Zach Furman, Logan Smith, Danny Halawi, Igor~V. Ostrovsky, Lev McKinney, Stella Biderman, and Jacob Steinhardt.
\newblock Eliciting latent predictions from transformers with the tuned lens.
\newblock \emph{ArXiv}, abs/2303.08112, 2023.

\bibitem[Bommasani et~al.(2021)Bommasani, Hudson, Adeli, Altman, Arora, von Arx, Bernstein, Bohg, Bosselut, Brunskill, et~al.]{bommasani2021opportunities}
Rishi Bommasani, Drew~A Hudson, Ehsan Adeli, Russ Altman, Simran Arora, Sydney von Arx, Michael~S Bernstein, Jeannette Bohg, Antoine Bosselut, Emma Brunskill, et~al.
\newblock On the opportunities and risks of foundation models.
\newblock \emph{arXiv preprint arXiv:2108.07258}, 2021.

\bibitem[Bubeck et~al.(2023)Bubeck, Chandrasekaran, Eldan, Gehrke, Horvitz, Kamar, Lee, Lee, Li, Lundberg, et~al.]{bubeck2023sparks}
S{\'e}bastien Bubeck, Varun Chandrasekaran, Ronen Eldan, Johannes Gehrke, Eric Horvitz, Ece Kamar, Peter Lee, Yin~Tat Lee, Yuanzhi Li, Scott Lundberg, et~al.
\newblock Sparks of artificial general intelligence: Early experiments with {GPT-4}.
\newblock \emph{arXiv preprint arXiv:2303.12712}, 2023.

\bibitem[Chang \& Bergen(2024)Chang and Bergen]{chang2024language}
Tyler~A Chang and Benjamin~K Bergen.
\newblock Language model behavior: A comprehensive survey.
\newblock \emph{Computational Linguistics}, 50\penalty0 (1):\penalty0 293--350, 2024.

\bibitem[Cunningham et~al.(2023)Cunningham, Ewart, Riggs, Huben, and Sharkey]{Cunningham2023SparseAF}
Hoagy Cunningham, Aidan Ewart, Logan Riggs, Robert Huben, and Lee Sharkey.
\newblock Sparse autoencoders find highly interpretable features in language models.
\newblock \emph{ArXiv}, abs/2309.08600, 2023.

\bibitem[Dou et~al.(2022)Dou, Forbes, Koncel-Kedziorski, Smith, and Choi]{dou2021scarecrow}
Yao Dou, Maxwell Forbes, Rik Koncel-Kedziorski, Noah~A. Smith, and Yejin Choi.
\newblock Is {GPT}-3 text indistinguishable from human text? {Scarecrow}: A framework for scrutinizing machine text.
\newblock In \emph{Proceedings of the 60th Annual Meeting of the Association for Computational Linguistics (Volume 1: Long Papers)}, pp.\  7250--7274, Dublin, Ireland, May 2022. Association for Computational Linguistics.
\newblock \doi{10.18653/v1/2022.acl-long.501}.

\bibitem[Ettinger et~al.(2016)Ettinger, Elgohary, and Resnik]{ettinger2016probing}
Allyson Ettinger, Ahmed Elgohary, and Philip Resnik.
\newblock Probing for semantic evidence of composition by means of simple classification tasks.
\newblock In \emph{Proceedings of the 1st Workshop on Evaluating Vector-Space Representations for NLP}, pp.\  134--139, 2016.

\bibitem[Geiger et~al.(2021)Geiger, Lu, Icard, and Potts]{geiger2021causal}
Atticus Geiger, Hanson Lu, Thomas Icard, and Christopher Potts.
\newblock Causal abstractions of neural networks.
\newblock \emph{Advances in Neural Information Processing Systems}, 34:\penalty0 9574--9586, 2021.

\bibitem[Geiger et~al.(2023)Geiger, Wu, Potts, Icard, and Goodman]{Geiger2023FindingAB}
Atticus Geiger, Zhengxuan Wu, Christopher Potts, Thomas~F. Icard, and Noah~D. Goodman.
\newblock Finding alignments between interpretable causal variables and distributed neural representations.
\newblock \emph{ArXiv}, abs/2303.02536, 2023.

\bibitem[Giulianelli et~al.(2018)Giulianelli, Harding, Mohnert, Hupkes, and Zuidema]{giulianelli2018hood}
Mario Giulianelli, Jack Harding, Florian Mohnert, Dieuwke Hupkes, and Willem Zuidema.
\newblock Under the hood: Using diagnostic classifiers to investigate and improve how language models track agreement information.
\newblock In Tal Linzen, Grzegorz Chrupa{\l}a, and Afra Alishahi (eds.), \emph{Proceedings of the 2018 {EMNLP} Workshop {B}lackbox{NLP}: Analyzing and Interpreting Neural Networks for {NLP}}, pp.\  240--248, Brussels, Belgium, November 2018. Association for Computational Linguistics.
\newblock \doi{10.18653/v1/W18-5426}.

\bibitem[Grattafiori et~al.(2024)Grattafiori, Dubey, Jauhri, Pandey, Kadian, Al-Dahle, Letman, and Mathur]{grattafiori2024llama3herdmodels}
Aaron Grattafiori, Abhimanyu Dubey, Abhinav Jauhri, Abhinav Pandey, Abhishek Kadian, Ahmad Al-Dahle, Aiesha Letman, and Akhil Mathur.
\newblock The llama 3 herd of models, 2024.

\bibitem[Griffiths et~al.(2024)Griffiths, Zhu, Grant, and Thomas~McCoy]{griffiths2024bayes}
Thomas~L Griffiths, Jian-Qiao Zhu, Erin Grant, and R~Thomas~McCoy.
\newblock Bayes in the age of intelligent machines.
\newblock \emph{Current Directions in Psychological Science}, 33\penalty0 (5):\penalty0 283--291, 2024.

\bibitem[Gurnee \& Tegmark(2024)Gurnee and Tegmark]{gurnee2024languagemodelsrepresentspace}
Wes Gurnee and Max Tegmark.
\newblock Language models represent space and time, 2024.

\bibitem[Gurnee et~al.(2024)Gurnee, Horsley, Guo, Kheirkhah, Sun, Hathaway, Nanda, and Bertsimas]{gurnee2024universal}
Wes Gurnee, Theo Horsley, Zifan~Carl Guo, Tara~Rezaei Kheirkhah, Qinyi Sun, Will Hathaway, Neel Nanda, and Dimitris Bertsimas.
\newblock Universal neurons in gpt2 language models.
\newblock \emph{arXiv preprint arXiv:2401.12181}, 2024.

\bibitem[Hewitt \& Liang(2019)Hewitt and Liang]{hewitt2019designing}
John Hewitt and Percy Liang.
\newblock Designing and interpreting probes with control tasks.
\newblock In Kentaro Inui, Jing Jiang, Vincent Ng, and Xiaojun Wan (eds.), \emph{Proceedings of the 2019 Conference on Empirical Methods in Natural Language Processing and the 9th International Joint Conference on Natural Language Processing (EMNLP-IJCNLP)}, pp.\  2733--2743, Hong Kong, China, November 2019. Association for Computational Linguistics.
\newblock \doi{10.18653/v1/D19-1275}.

\bibitem[Hu et~al.(2021)Hu, Shen, Wallis, Allen-Zhu, Li, Wang, and Chen]{Hu2021LoRALA}
J.~Edward Hu, Yelong Shen, Phillip Wallis, Zeyuan Allen-Zhu, Yuanzhi Li, Shean Wang, and Weizhu Chen.
\newblock Lora: Low-rank adaptation of large language models.
\newblock \emph{ArXiv}, abs/2106.09685, 2021.

\bibitem[Hu \& Collier(2024)Hu and Collier]{Hu2024QuantifyingTP}
Tiancheng Hu and Nigel Collier.
\newblock Quantifying the persona effect in llm simulations.
\newblock \emph{ArXiv}, abs/2402.10811, 2024.

\bibitem[Hupkes et~al.(2018)Hupkes, Veldhoen, and Zuidema]{hupkes2018visualisation}
Dieuwke Hupkes, Sara Veldhoen, and Willem Zuidema.
\newblock Visualisation and `diagnostic classifiers' reveal how recurrent and recursive neural networks process hierarchical structure.
\newblock \emph{Journal of Artificial Intelligence Research}, 61:\penalty0 907--926, 2018.

\bibitem[Kadavath et~al.(2022)Kadavath, Conerly, Askell, Henighan, Drain, Perez, Schiefer, Hatfield-Dodds, DasSarma, Tran-Johnson, Johnston, El-Showk, Jones, Elhage, Hume, Chen, Bai, Bowman, Fort, Ganguli, Hernandez, Jacobson, Kernion, Kravec, Lovitt, Ndousse, Olsson, Ringer, Amodei, Brown, Clark, Joseph, Mann, McCandlish, Olah, and Kaplan]{kadavath2022language}
Saurav Kadavath, Tom Conerly, Amanda Askell, Tom Henighan, Dawn Drain, Ethan Perez, Nicholas Schiefer, Zac Hatfield-Dodds, Nova DasSarma, Eli Tran-Johnson, Scott Johnston, Sheer El-Showk, Andy Jones, Nelson Elhage, Tristan Hume, Anna Chen, Yuntao Bai, Sam Bowman, Stanislav Fort, Deep Ganguli, Danny Hernandez, Josh Jacobson, Jackson Kernion, Shauna Kravec, Liane Lovitt, Kamal Ndousse, Catherine Olsson, Sam Ringer, Dario Amodei, Tom Brown, Jack Clark, Nicholas Joseph, Ben Mann, Sam McCandlish, Chris Olah, and Jared Kaplan.
\newblock Language models (mostly) know what they know.
\newblock \emph{arXiv preprint arXiv:2207.05221}, 2022.

\bibitem[Katz \& Belinkov(2023)Katz and Belinkov]{katz-belinkov-2023-visit}
Shahar Katz and Yonatan Belinkov.
\newblock {VISIT}: Visualizing and interpreting the semantic information flow of transformers.
\newblock In Houda Bouamor, Juan Pino, and Kalika Bali (eds.), \emph{Findings of the Association for Computational Linguistics: EMNLP 2023}, pp.\  14094--14113, Singapore, December 2023. Association for Computational Linguistics.
\newblock \doi{10.18653/v1/2023.findings-emnlp.939}.

\bibitem[Kojima et~al.(2022)Kojima, Gu, Reid, Matsuo, and Iwasawa]{kojima2022large}
Takeshi Kojima, Shixiang~Shane Gu, Machel Reid, Yutaka Matsuo, and Yusuke Iwasawa.
\newblock Large language models are zero-shot reasoners.
\newblock \emph{Advances in Neural Information Processing Systems}, 35:\penalty0 22199--22213, 2022.

\bibitem[Kotek et~al.(2023)Kotek, Dockum, and Sun]{kotek2023gender}
Hadas Kotek, Rikker Dockum, and David Sun.
\newblock Gender bias and stereotypes in large language models.
\newblock In \emph{Proceedings of the ACM collective intelligence conference}, pp.\  12--24, 2023.

\bibitem[Lin et~al.(2022)Lin, Hilton, and Evans]{lin-etal-2022-truthfulqa}
Stephanie Lin, Jacob Hilton, and Owain Evans.
\newblock {T}ruthful{QA}: Measuring how models mimic human falsehoods.
\newblock In Smaranda Muresan, Preslav Nakov, and Aline Villavicencio (eds.), \emph{Proceedings of the 60th Annual Meeting of the Association for Computational Linguistics (Volume 1: Long Papers)}, pp.\  3214--3252, Dublin, Ireland, May 2022. Association for Computational Linguistics.
\newblock \doi{10.18653/v1/2022.acl-long.229}.

\bibitem[Liu et~al.(2022)Liu, Zhang, Brockett, Mao, Sui, Chen, and Dolan]{liu2022token}
Tianyu Liu, Yizhe Zhang, Chris Brockett, Yi~Mao, Zhifang Sui, Weizhu Chen, and Bill Dolan.
\newblock A token-level reference-free hallucination detection benchmark for free-form text generation.
\newblock In Smaranda Muresan, Preslav Nakov, and Aline Villavicencio (eds.), \emph{Proceedings of the 60th Annual Meeting of the Association for Computational Linguistics (Volume 1: Long Papers)}, pp.\  6723--6737, Dublin, Ireland, May 2022. Association for Computational Linguistics.
\newblock \doi{10.18653/v1/2022.acl-long.464}.

\bibitem[McCoy et~al.(2023)McCoy, Yao, Friedman, Hardy, and Griffiths]{mccoy2023embersautoregressionunderstandinglarge}
R.~Thomas McCoy, Shunyu Yao, Dan Friedman, Matthew Hardy, and Thomas~L. Griffiths.
\newblock Embers of autoregression: Understanding large language models through the problem they are trained to solve, 2023.

\bibitem[Minder et~al.(2025)Minder, Du, Stoehr, Monea, Wendler, West, and Cotterell]{minder2025controllablecontextsensitivityknob}
Julian Minder, Kevin Du, Niklas Stoehr, Giovanni Monea, Chris Wendler, Robert West, and Ryan Cotterell.
\newblock Controllable context sensitivity and the knob behind it, 2025.

\bibitem[Mirzadeh et~al.(2024)Mirzadeh, Alizadeh, Shahrokhi, Tuzel, Bengio, and Farajtabar]{mirzadeh2024gsm}
Iman Mirzadeh, Keivan Alizadeh, Hooman Shahrokhi, Oncel Tuzel, Samy Bengio, and Mehrdad Farajtabar.
\newblock {GSM}-{S}ymbolic: Understanding the limitations of mathematical reasoning in large language models.
\newblock \emph{arXiv preprint arXiv:2410.05229}, 2024.

\bibitem[Nanda et~al.(2023)Nanda, Lee, and Wattenberg]{nanda-etal-2023-emergent}
Neel Nanda, Andrew Lee, and Martin Wattenberg.
\newblock Emergent linear representations in world models of self-supervised sequence models.
\newblock In Yonatan Belinkov, Sophie Hao, Jaap Jumelet, Najoung Kim, Arya McCarthy, and Hosein Mohebbi (eds.), \emph{Proceedings of the 6th BlackboxNLP Workshop: Analyzing and Interpreting Neural Networks for NLP}, Singapore, December 2023. Association for Computational Linguistics.

\bibitem[Nielsen et~al.(2025)Nielsen, Macocco, and Baroni]{nielsen2025predictionhubscontextinformedfrequent}
Beatrix M.~G. Nielsen, Iuri Macocco, and Marco Baroni.
\newblock Prediction hubs are context-informed frequent tokens in llms, 2025.

\bibitem[Orgad et~al.(2024)Orgad, Toker, Gekhman, Reichart, Szpektor, Kotek, and Belinkov]{orgad2024llmsknowshowintrinsic}
Hadas Orgad, Michael Toker, Zorik Gekhman, Roi Reichart, Idan Szpektor, Hadas Kotek, and Yonatan Belinkov.
\newblock Llms know more than they show: On the intrinsic representation of llm hallucinations, 2024.

\bibitem[Prabhakar et~al.(2024)Prabhakar, Griffiths, and McCoy]{Prabhakar2024DecipheringTF}
Akshara Prabhakar, Thomas~L. Griffiths, and R.~Thomas McCoy.
\newblock Deciphering the factors influencing the efficacy of chain-of-thought: Probability, memorization, and noisy reasoning.
\newblock \emph{ArXiv}, abs/2407.01687, 2024.

\bibitem[Radford et~al.(2019)Radford, Wu, Child, Luan, Amodei, and Sutskever]{radford2019language}
Alec Radford, Jeffrey Wu, Rewon Child, David Luan, Dario Amodei, and Ilya Sutskever.
\newblock Language models are unsupervised multitask learners.
\newblock \emph{OpenAI Blog}, 2019.

\bibitem[Raffel et~al.(2023)Raffel, Shazeer, Roberts, Lee, Narang, Matena, Zhou, Li, and Liu]{raffel2023exploringlimitstransferlearning}
Colin Raffel, Noam Shazeer, Adam Roberts, Katherine Lee, Sharan Narang, Michael Matena, Yanqi Zhou, Wei Li, and Peter~J. Liu.
\newblock Exploring the limits of transfer learning with a unified text-to-text transformer, 2023.

\bibitem[Razeghi et~al.(2022)Razeghi, Logan~IV, Gardner, and Singh]{razeghi2022impact}
Yasaman Razeghi, Robert~L Logan~IV, Matt Gardner, and Sameer Singh.
\newblock Impact of pretraining term frequencies on few-shot numerical reasoning.
\newblock In Yoav Goldberg, Zornitsa Kozareva, and Yue Zhang (eds.), \emph{Findings of the Association for Computational Linguistics: EMNLP 2022}, pp.\  840--854, Abu Dhabi, United Arab Emirates, December 2022. Association for Computational Linguistics.
\newblock \doi{10.18653/v1/2022.findings-emnlp.59}.

\bibitem[Rimsky et~al.(2023)Rimsky, Gabrieli, Schulz, Tong, Hubinger, and Turner]{Rimsky2023SteeringL2}
Nina Rimsky, Nick Gabrieli, Julian Schulz, Meg Tong, Evan Hubinger, and Alexander~Matt Turner.
\newblock Steering llama 2 via contrastive activation addition.
\newblock \emph{ArXiv}, abs/2312.06681, 2023.

\bibitem[Salinas \& Morstatter(2024)Salinas and Morstatter]{salinas2024butterfly}
Abel Salinas and Fred Morstatter.
\newblock The butterfly effect of altering prompts: How small changes and jailbreaks affect large language model performance.
\newblock \emph{arXiv preprint arXiv:2401.03729}, 2024.

\bibitem[Smolensky(1986)]{smolensky1986neural}
Paul Smolensky.
\newblock Neural and conceptual interpretation of pdp models.
\newblock \emph{Parallel distributed processing: Explorations in the microstructure of cognition}, 2:\penalty0 390--431, 1986.

\bibitem[Soulos et~al.(2020)Soulos, McCoy, Linzen, and Smolensky]{soulos2020discovering}
Paul Soulos, R.~Thomas McCoy, Tal Linzen, and Paul Smolensky.
\newblock Discovering the compositional structure of vector representations with role learning networks.
\newblock In Afra Alishahi, Yonatan Belinkov, Grzegorz Chrupa{\l}a, Dieuwke Hupkes, Yuval Pinter, and Hassan Sajjad (eds.), \emph{Proceedings of the Third BlackboxNLP Workshop on Analyzing and Interpreting Neural Networks for NLP}, pp.\  238--254, Online, November 2020. Association for Computational Linguistics.
\newblock \doi{10.18653/v1/2020.blackboxnlp-1.23}.

\bibitem[Stechly et~al.(2024)Stechly, Valmeekam, and Kambhampati]{stechly2024chain}
Kaya Stechly, Karthik Valmeekam, and Subbarao Kambhampati.
\newblock Chain of thoughtlessness? an analysis of cot in planning.
\newblock In \emph{The Thirty-eighth Annual Conference on Neural Information Processing Systems}, 2024.

\bibitem[Stolfo et~al.(2024)Stolfo, Wu, Gurnee, Belinkov, Song, Sachan, and Nanda]{stolfo2024confidenceregulationneuronslanguage}
Alessandro Stolfo, Ben Wu, Wes Gurnee, Yonatan Belinkov, Xingyi Song, Mrinmaya Sachan, and Neel Nanda.
\newblock Confidence regulation neurons in language models, 2024.

\bibitem[Templeton(2024)]{templeton2024scaling}
Adly Templeton.
\newblock \emph{Scaling monosemanticity: Extracting interpretable features from claude 3 sonnet}.
\newblock Anthropic, 2024.

\bibitem[Tian et~al.(2023)Tian, Mitchell, Zhou, Sharma, Rafailov, Yao, Finn, and Manning]{tian-etal-2023-just}
Katherine Tian, Eric Mitchell, Allan Zhou, Archit Sharma, Rafael Rafailov, Huaxiu Yao, Chelsea Finn, and Christopher Manning.
\newblock Just ask for calibration: Strategies for eliciting calibrated confidence scores from language models fine-tuned with human feedback.
\newblock In Houda Bouamor, Juan Pino, and Kalika Bali (eds.), \emph{Proceedings of the 2023 Conference on Empirical Methods in Natural Language Processing}, pp.\  5433--5442, Singapore, December 2023. Association for Computational Linguistics.
\newblock \doi{10.18653/v1/2023.emnlp-main.330}.

\bibitem[torchtune maintainers \& contributors(2024)torchtune maintainers and contributors]{torchtune}
torchtune maintainers and contributors.
\newblock torchtune: Pytorch's post-training library, 2024.
\newblock URL \url{https//github.com/pytorch/torchtune}.

\bibitem[Vaswani et~al.(2017)Vaswani, Shazeer, Parmar, Uszkoreit, Jones, Gomez, Kaiser, and Polosukhin]{Vaswani+2017}
Ashish Vaswani, Noam Shazeer, Niki Parmar, Jakob Uszkoreit, Llion Jones, Aidan~N Gomez, \L~ukasz Kaiser, and Illia Polosukhin.
\newblock Attention is all you need.
\newblock In \emph{Advances in Neural Information Processing Systems}, volume~30. Curran Associates, Inc., 2017.

\bibitem[Venkit et~al.(2024)Venkit, Chakravorti, Gupta, Biggs, Srinath, Goswami, Rajtmajer, and Wilson]{venkit2024audit}
Pranav~Narayanan Venkit, Tatiana Chakravorti, Vipul Gupta, Heidi Biggs, Mukund Srinath, Koustava Goswami, Sarah Rajtmajer, and Shomir Wilson.
\newblock An audit on the perspectives and challenges of hallucinations in nlp.
\newblock \emph{arXiv preprint arXiv:2404.07461}, 2024.

\bibitem[Wei et~al.(2022)Wei, Wang, Schuurmans, Bosma, Chi, Xia, Le, and Zhou]{Wei2022ChainOT}
Jason Wei, Xuezhi Wang, Dale Schuurmans, Maarten Bosma, Ed~H. Chi, F.~Xia, Quoc Le, and Denny Zhou.
\newblock Chain of thought prompting elicits reasoning in large language models.
\newblock \emph{ArXiv}, abs/2201.11903, 2022.

\bibitem[Wieting \& Kiela(2019)Wieting and Kiela]{wieting2018no}
John Wieting and Douwe Kiela.
\newblock No training required: Exploring random encoders for sentence classification.
\newblock In \emph{International Conference on Learning Representations}, 2019.

\bibitem[Wu et~al.(2024{\natexlab{a}})Wu, Qiu, Ross, Aky{\"u}rek, Chen, Wang, Kim, Andreas, and Kim]{wu2024reasoning}
Zhaofeng Wu, Linlu Qiu, Alexis Ross, Ekin Aky{\"u}rek, Boyuan Chen, Bailin Wang, Najoung Kim, Jacob Andreas, and Yoon Kim.
\newblock Reasoning or reciting? exploring the capabilities and limitations of language models through counterfactual tasks.
\newblock In \emph{Proceedings of the 2024 Conference of the North American Chapter of the Association for Computational Linguistics: Human Language Technologies (Volume 1: Long Papers)}, pp.\  1819--1862, 2024{\natexlab{a}}.

\bibitem[Wu et~al.(2024{\natexlab{b}})Wu, Arora, Wang, Geiger, Jurafsky, Manning, and Potts]{Wu2024ReFTRF}
Zhengxuan Wu, Aryaman Arora, Zheng Wang, Atticus Geiger, Daniel Jurafsky, Christopher~D. Manning, and Christopher Potts.
\newblock Reft: Representation finetuning for language models.
\newblock \emph{ArXiv}, abs/2404.03592, 2024{\natexlab{b}}.

\bibitem[Yao et~al.(2023)Yao, Yu, Zhao, Shafran, Griffiths, Cao, and Narasimhan]{Yao2023TreeOT}
Shunyu Yao, Dian Yu, Jeffrey Zhao, Izhak Shafran, Thomas~L. Griffiths, Yuan Cao, and Karthik Narasimhan.
\newblock Tree of thoughts: Deliberate problem solving with large language models.
\newblock \emph{ArXiv}, abs/2305.10601, 2023.

\bibitem[Zhu \& Griffiths(2024)Zhu and Griffiths]{zhu2024incoherentprobabilityjudgmentslarge}
Jian-Qiao Zhu and Thomas~L. Griffiths.
\newblock Incoherent probability judgments in large language models, 2024.

\end{thebibliography}
\bibliographystyle{colm2025_conference}

\newpage
\appendix
\section{Failed Techniques to Reduce Reliance on Prior}
\label{sec:appendix-techniques}
In this section we share some of the techniques we tried to reduce the models reliance on the prior, but did not effectively do so. 

\subsection{Context vs.\ prior steering}
Prior work has shown that there exists a one-dimensional subspace responsible for the language model attending to its prior or the context~\citep{minder2025controllablecontextsensitivityknob}. This work was motivated by the observation that often times we want to state facts in the context that conflict with the models prior; for example, the capital of France becomes London we want an in-context update to have the model answer London when asked what is the capital of France. 

We hypothesized that steering on this dimension to rely on the context more than the prior would result in comparable gains to probing, since the ability to overwrite what it ``expects'' the answer to be would be gone. 

To account for this, we took the approach from the original paper (see Table~\ref{tab:app:params} for hyperparameters used) and re-ran the analysis using their exact setup. We found that this form of steering had a minimal impact on performance as illustrated in Table~\ref{tab:app:performance}. We also tested different scaling parameters and found that they only dropped accuracy. 

\begin{table}[t!]
    \centering
    \begin{tabular}{c|c|c}
        Setting & Scaling factor& Layer \\ 
        \midrule
         Context vs.\ prior steering& -1 & 16 \\
        Unigram prior removal& -10 & -1 \\ 
    \end{tabular}
    \caption{Hyperparameters for using the context vs.\ prior and unigram prior removal steering.}
    \label{tab:app:params}
\end{table}

\subsection{Ablating the unigram prior direction}
Past work has shown that LLMs have mechanisms for handling uncertainty~\citep{stolfo2024confidenceregulationneuronslanguage}. One of these is the so-called unigram neuron that pushes a models response towards 

We take motivation from this approach an ablate out the unigram direction from the model. To do this we first projected the token-frequency direction as calculated by a sample of the C4 dataset~\citep{raffel2023exploringlimitstransferlearning} and ablated on this direction. In other words, we no longer let the LLM write into the direction of the unigram distribution thereby limiting the effect of the unigram frequency neurons. 

Again, we found that this approach didn't work as evidenced in Table~\ref{tab:app:performance}.

\begin{table}[]
    \centering
    \begin{tabular}{c|c|c}
        & \textbf{Counting} & \textbf{Acronym} \\
        Context vs.\ prior steering &$0.0953$ & $0.0390$  \\
        Unigram prior removal  & $0.198$ & $0.067$ 
    \end{tabular}
    \caption{Performance using the context vs.\ prior and unigram prior removal steering.}
    \label{tab:app:performance}
\end{table}


\section{Prompt - Ground Truth Response Examples}
\label{sec:appendix-prompts}

\begin{figure*}[t]
    \centering
    \subfigure[Multiplication.]{
    \includegraphics[width=0.8\linewidth]{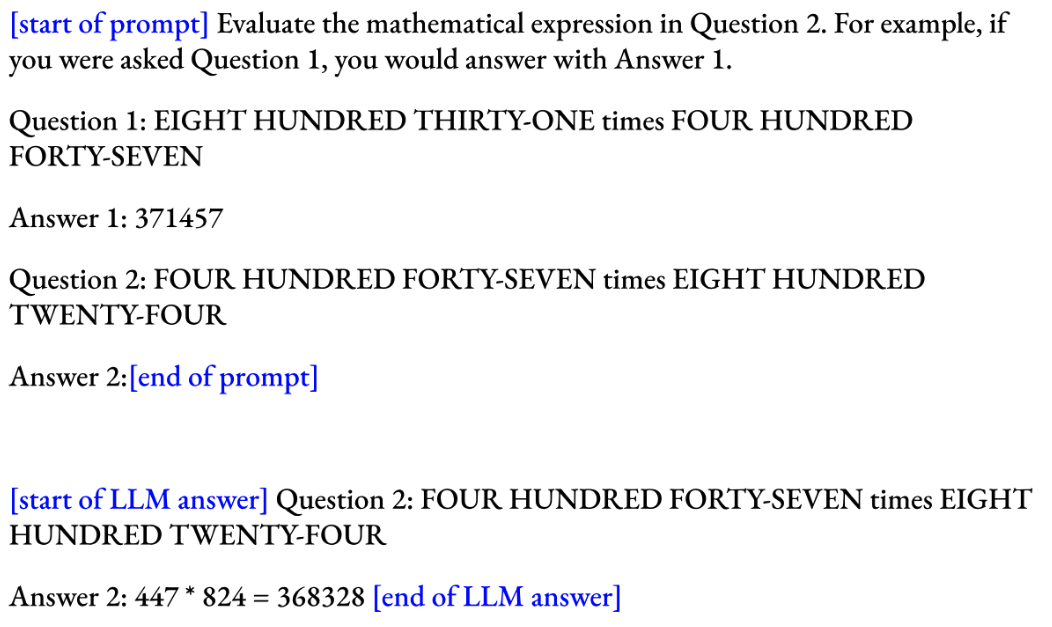}
    }
    \hfill
    \subfigure[Produce letters.]{
    \includegraphics[width=0.8\linewidth]{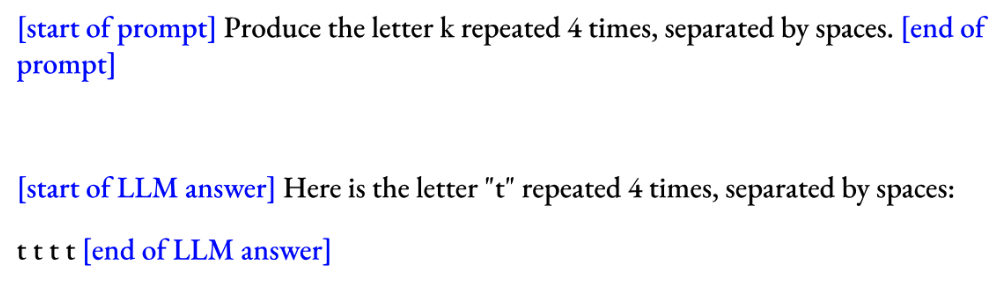}
    }
    \hfill
    \subfigure[Counting]{
    \includegraphics[width=0.8\linewidth]{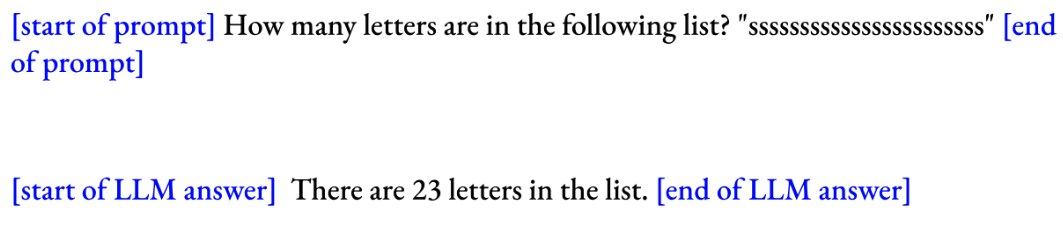}
    }
    \hfill
    \subfigure[Acronyms.]{
    \includegraphics[width=0.8\linewidth]{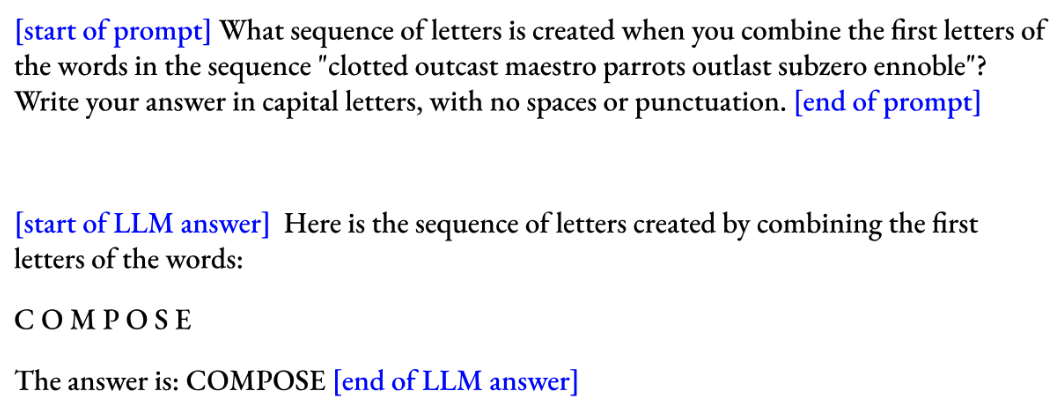}
    }
    \caption{Prompt and ground truth examples of each task (shift-cipher example is included Figure \ref{fig:shift-cipher-example}). On multiplication, we tried arabic numbers, English caps, English lowercases in the prompt, but English caps achieved the highest base accuracy, so this is the version we use in our experiments, which is also the most conservative version.}
    \label{fig:prompt-example}
\end{figure*}

One example prompt and a ground truth response example for each task is shown in Figure \ref{fig:prompt-example}.

\section{Implementation Details}
\label{sec:appendix-implementation}

We use the torchtune package for finetuning \citep{torchtune}. All experiments are trained for 50 epochs, use learning-rate $=10^{-4}$, weight-decay $=0.01$, batch-size$=2$, gradient accumulation steps $=8$, LoRA rank $=8$, LoRA alpha $=16$, and LoRA is applied to attention modules and each MLP. Each model finetuning uses one Nvidia A100 GPU.

\end{document}